\documentclass{acmart}

\usepackage{booktabs} 
\usepackage{xspace}
\usepackage{multicol}
\usepackage{keycommand}
\usepackage{tikz}
\usetikzlibrary{automata,shapes,arrows,shadows,calc,trees,decorations.pathreplacing,patterns}
\usepackage{pgfplots}
\usepackage{pifont}
\usepackage{enumerate}
\usepackage{multirow}
\usepackage{array}
\usepackage{verbatim}
\usepackage{enumitem}

\newcommand{\R}{\ensuremath{\mathcal{R}}\xspace}
\newcommand{\F}{\ensuremath{\mathcal{F}}\xspace}
\newcommand{\T}{\ensuremath{\mathcal{T}}\xspace}
\newcommand{\V}{\ensuremath{\mathcal{V}}\xspace}

\newcommand{\C}{\ensuremath{\mathcal{C}}\xspace}
\renewcommand{\S}{\ensuremath{\Sigma}\xspace}
\renewcommand{\P}{\ensuremath{\mathcal{P}}\xspace}

\newcommand{\rewrite}{\ensuremath{\ \texttt{=}\ }}
\newcommand{\haea}{\textbf{\textsc{HaEa}}\xspace}
\newcommand{\Var}{\ensuremath{\mathcal{V}ar}\xspace}

\newkeycommand\barchart[scale=0.1,experiments=100,success=0,persuccess=0,minlength=1]{
\raisebox{-2pt}{
\begin{tikzpicture}[xscale=\commandkey{scale},yscale=1.0pt,line width=0.8pt]
\draw[draw=white] (0.0,0.0) rectangle (\commandkey{experiments},0.35);
\draw[pattern=north east lines,fill,draw] (0.0,0.0) rectangle (\commandkey{success},0.3);
\draw[draw] (0.0,0.0) rectangle (\commandkey{experiments},0.3);
\draw (\commandkey{experiments},0.15) node[right=1pt,inner sep=0pt] {\tiny\textsf{\commandkey{persuccess}\%}};
\end{tikzpicture}}}

\newkeycommand\bigboxwhiskers[title=Title,numberbox=1,lowerinnerfence=0,min=0,Q1=0,median=0,Q3=0,max=0,upperinnerfence=0,mean=0]{
\begin{axis}
[x=8mm,y=0.4mm,xmin=-0.5,xmax=7,ymin=-12,ymax=106,axis lines=none]
\def\abscissa{3 * \commandkey{numberbox} - 2};
\begin{scope}[line width=1.0pt]
\draw (axis cs:\abscissa,\commandkey{Q1}) rectangle (axis cs:\abscissa + 2,\commandkey{Q3});
\ifdim \commandkey{lowerinnerfence}pt > \commandkey{min}pt 
\draw (axis cs:\abscissa + 0.25,\commandkey{lowerinnerfence}) -- (axis cs:\abscissa + 1.75,\commandkey{lowerinnerfence});
\draw (axis cs:\abscissa + 1,\commandkey{lowerinnerfence}) -- (axis cs:\abscissa + 1,\commandkey{Q1});
\draw[dashed] (axis cs:\abscissa + 0.5,\commandkey{min}) -- (axis cs:\abscissa + 1.5,\commandkey{min});
\draw[dashed] (axis cs:\abscissa + 1,\commandkey{min}) -- (axis cs:\abscissa + 1,\commandkey{lowerinnerfence});
\else
\draw (axis cs:\abscissa + 0.25,\commandkey{min}) -- (axis cs:\abscissa + 1.75,\commandkey{min});
\draw (axis cs:\abscissa + 1,\commandkey{min}) -- (axis cs:\abscissa + 1,\commandkey{Q1});
\fi
\ifdim \commandkey{upperinnerfence}pt < \commandkey{max}pt 
\draw (axis cs:\abscissa + 0.25,\commandkey{upperinnerfence}) -- (axis cs:\abscissa + 1.75,\commandkey{upperinnerfence});
\draw (axis cs:\abscissa + 1,\commandkey{upperinnerfence}) -- (axis cs:\abscissa + 1,\commandkey{Q3});
\draw[dashed] (axis cs:\abscissa + 0.5,\commandkey{max}) -- (axis cs: \abscissa + 1.5,\commandkey{max});
\draw[dashed] (axis cs:\abscissa + 1,\commandkey{max}) -- (axis cs:\abscissa + 1,\commandkey{upperinnerfence});
\else
\draw (axis cs:\abscissa + 0.25,\commandkey{max}) -- (axis cs:\abscissa + 1.75,\commandkey{max});
\draw (axis cs:\abscissa + 1,\commandkey{max}) -- (axis cs:\abscissa + 1,\commandkey{Q3});
\fi
\draw[line width=3pt,dotted] (axis cs:\abscissa,\commandkey{median}) -- (axis cs:\abscissa + 2,\commandkey{median});
\draw (axis cs:\abscissa + 1,\commandkey{mean}) node {\scriptsize$\boldsymbol{\star}$};
\draw (axis cs:\abscissa + 1,-10) node {\scriptsize\commandkey{title}};
\end{scope}
\end{axis}
}

\theoremstyle{definition}

\begin{document}
\title{Obtaining Basic Algebra Formulas with Genetic Programming and Functional Rewriting}

\author{Edwin Camilo Cubides}
\affiliation{%
  \institution{Universidad Nacional de Colombia}
  \streetaddress{Av. Carrera 30 No. 45-03}
  \city{Bogot\'a} 
  \country{Colombia} 
}
\email{eccubidesg@unal.edu.co}

\author{Jonatan Gom\'ez}
\affiliation{%
  \institution{Universidad Nacional de Colombia}
  \streetaddress{Av. Carrera 30 No. 45-03}
  \city{Bogot\'a} 
  \country{Colombia} 
}
\email{jgomezpe@unal.edu.co}

\renewcommand{\shortauthors}{E. C. Cubides \& J. Gom\'ez}

\begin{abstract}
In this paper we develop a set of genetic programming operators and an initialization population process based on concepts of functional programming rewriting for boosting inductive genetic programming. Such genetic operators are used with in a hybrid adaptive evolutionary algorithm that evolves operator rates at the same time it evolves the solution. Solutions are represented using recursive functions where genome is encoded as an ordered list of trees and phenotype is written in a simple functional programming language that uses rewriting as operational semantic (computational model). The fitness is the number of examples successfully deduced over the cardinal of the set of examples. Parents are selected following a tournament selection mechanism and the next population is obtained following a steady state strategy. The evolutionary process can use some previous functions (programs) induced as background knowledge. We compare the performance of our technique in a set of hard problems (for classical genetic programming). In particular, we take as test-bed the problem of obtaining equivalent algebraic expressions of some notable products (such as square of a binomial, and cube of a binomial), and the recursive formulas of sum of the first $n$ and squares of the first $n$ natural numbers.
\end{abstract}

%
%


\terms{Genetic Programming, Inductive Learning}



\maketitle

\section{Introduction}

Nowadays the Machine Learning area has been deeply and extensively studied \cite{Mitchell97}. The Machine Learning is the Artificial Intelligence area that use inductive learning and deals with the problem of generating a model or hypothesis (learn a task) to approximate a function well over a sufficiently large set of training examples, and automatically to improve this model to explain other unobserved examples if it is enough robust \cite{MichieSpiegelhalterTaylor94,Mitchell97}. The majority of problem that Machine Learning tries to solve are difficult, mainly due to their complex nature and their representation. 

In particular, a problem of this kind is the automatic programming problem, which can be defined as the automatic process to obtain a computer program with some mechanism and where humans intervention is low \cite{IgwePillay13}. It is very complicated because its representation depends on the target programming language. Mainly, there are two types of automatic programming: \emph{Deductive} where the program is generated from a high-level description, and \emph{Inductive} where the program is generated from a set of examples (pairs input/output).

Genetic Programming (GP) \cite{Koza92,Koza94,Koza99,Koza03} is an inductive programming technique and a type of \emph{soft computing technique}. In GP, programs are represented using some data structure (usually trees). Basically, programs are randomly generated and evolved by applying some genetic operators 

Inductive Programming (IP) is an inductive technique \cite{Muggleton91,Muggleton92,Plotkin70,Plotkin71,Shapiro91} and a type of \emph{hard computing technique}. In IP programs are represented using some logic language.  The inductive method is very interesting since the theoretical point of view, but is unacceptable since the computational point of view, because this method grows exponentially in time and space \cite{HernandezOrallo2011InverseNF}. Genetic Inductive Programming (GIP) \cite{GILP1997Tveit,Varsek93,WongLeung94} is an inductive Machine Learning technique that uses the formal representation and non deterministic search, and its goal is to combine GP and IP to reduce the search space and the required time to run the algorithm. 

Given dataset of examples in form of pair input/output, it is interesting describes those examples as a function (expression) that represent the set in a compact format, as application of techniques described previously we present some examples of concrete cases of arithmetic pairs and use an evolutionary algorithm proposed for finding solution as abstract algebraic expressions that describe all examples as an unify theory, the solution found for the dataset presented could be associated to notable product of elemental algebra. 

This paper is organized as follows. Section~2 presents the basic concepts and theoretic fundaments about term rewriting system and functional programming. Section~3 describes the hybrid adaptive evolutionary algorithm that adapts the operator rates while it is solving the optimization problem. Section~4 the evolutionary algorithm proposed to evolve functional programs is described. Section~5 explain the configuration of the experiments that was done and the result are analyzed. Finally, we presented the conclusions obtained from the experiments done.      

\section{Term Rewriting System and\\Functional Programming}

Concepts as Equational Logic and Term Rewriting Systems are necessary to understand the theory behind of the learning induction of programs written on a declarative language. A short introduction to these concepts is done in this section (a full explanation can be found in \cite{BaaderNipkow98}).

Let \V be a countable set of \emph{variables}. Let \S denotes a non-empty, finite set of \emph{function symbols} or \emph{signature}, each of which has a fixed associated $arity$, it is defined as a mapping from \S into $\mathbb{N}$ such that $arity(f) = n$, where $n$ is the number of parameters of $f$, we often write $f/n$ to denote that $arity(f) = n$. If $arity(f) = 0$ then the function symbol $f$ is called a \emph{constant} (it is supposed that \S contains at least one constant). The expression $\T(\S,\V)$ denotes the set of terms built from \S and \V. The set of variables occurring in a term $t$ is denoted $\Var(t)$. A term $t$ is a \emph{ground term} if $Var(t) = \varnothing$. A \emph{fresh} variable is a variable that appears nowhere else.

An \emph{occurrence} $u$ in a term $t$ is represented by a sequence of natural numbers. These sequences belong to the language given by the regular expression $\{\Lambda\} \cup \{n(.m)^* : n,m \in \mathbb{N}^+\}$. $O(t)$ denotes the set of occurrences and $\overline{O}(t)$ denotes the set of non-variables occurrences of $t$. $t|_u$ denotes the \emph{subterm} of $t$ in the occurrence $u$. $t[u]$ denotes the leftmost symbol of the subterm of $t$ in the occurrence $u$. $t[s]_u$ denotes the \emph{replacement} of the subterm of $t$ in the occurrence $u$ by the term $s$.

A \emph{substitution} $\sigma$ is defined as a mapping from the set of variables \V into the set of terms $\T(\S,\V)$. If it is necessary, a substitution can be extended as an homomorphism from $\T(\S,\V)$ into $\T(\S,\V)$, and it is applied only on the variables of the term. A substitution is usually represented by $\sigma = \{x_1/t_1, \ldots, x_n/t_n\}$ where $\sigma(x_i) = t_i \neq x_i$, $i=1,\ldots,n$. The \emph{identity} substitution $id$ is the empty substitution $\{\ \}$. A \emph{ground} substitution is a substitution where all $t_i$ are ground terms. A term $s$ is an instance of a term $t$ if there exist a substitution $\sigma$ such that $s = \sigma(t)$. The composition of two substitutions $\delta$ and $\sigma$ is denoted by $(\delta \circ \sigma)(t)$ and it is done as the composition between functions, i.e. $(\delta \circ \sigma)(t) = \delta(\sigma(t))$. The expression $\sigma \leq \theta$ means that there exist a substitution $\delta$ such that $\theta = \delta \circ \sigma$. A substitution $\sigma$ is an \emph{unifier} of two term $t$ and $s$ if $\sigma(t) = \sigma(s)$. $\sigma$ is a \emph{most general unifier} (mgu) if it is an unifier and $\sigma \leq \theta$ for any another unifier $\theta$.

A \emph{equation} or \emph{rewriting rule} is an expression $l \rewrite r$, where $l$ and $r$ are terms, $l$ is called the \emph{left-hand side} (lhs), of the equation, and $r$ is the \emph{right-hand side} (rhs). A \emph{Term Rewriting System} (TRS) is a finite set of rewriting rules. A \emph{functional program} \P is a TRS in which every rule satisfies $\Var(r) \subseteq \Var(l)$. Given an equation, an \emph{orphan} variable $x$ in the equation is a variable such that $x \in \Var(r)$ and $x \notin \Var(l)$.

For TRS \R , $r \ll \R$ denotes that $r$ is a new variant of a rule in \R such that $r$ contains only fresh variables, i.e. contains no variable previously met during computation. Given a TRS \R, we assume that the signature \S is partitioned into two disjoint sets $\S = \C \uplus \F$, where $\F = \{f : f(t_1,\dots,t_n) \rewrite r \in \R\}$ and $\C = \S \setminus \F$. Symbols in \C are called \emph{constructors} and symbols in \F are called \emph{defined functions}. The elements of $\T(\C,\V)$ are called \emph{constructor terms}. A \emph{constructor substitution} $\sigma = \{x_1/t_1, \ldots, x_n/t_n\}$ is a substitution such that each $t_i$, $i = 1,\ldots,n$ is a constructor term. A term is linear if it does not contains multiple occurrences of the same variable. A TRS is left-linear if the left-hand sides of all rules are linear terms. A \emph{pattern} is a term of the form $f(c_1,\ldots,c_n)$ where $f \in \F$ and $c_1,\ldots,c_n$ are constructor terms. We say that a TRS is \emph{construct-based} (CB) is the left-hand sides of \R are patterns.

Given a TRS $\R$, $t \xrightarrow{u,l \rewrite r,\sigma}_\R s$ is a \emph{rewrite step} if there exist an occurrence $u$ of $t$, a rule $l\rewrite r \in \R$ and a substitution $\sigma$ such that $t|_u = \sigma(l)$ and $s = t[\sigma(r)]_u$. A term $t$ is said to be in \emph{normal form} with respect to (w.r.t) \R if there exists not a term $s$ such that $t \xrightarrow{u,l \rewrite r,\sigma}_\R s$. An equation $t \rewrite s$ is \emph{normalized} w.r.t \R if $t$ and $s$ are in normal form. \R is said to be \emph{terminating} or is \emph{noetheriano} if there exists not an infinite rewrite steps chain $t_1 \to_\R t_2 \to_\R t_3 \to_\R \cdots$. \R is said to be \emph{confluent} if for all $t_1,t_2,t_3 \in \T(\F,\V)$ with $t_1 \to_\R^* t_2$ and $t_1 \to_\R^* t_3$, then there exists $t \in \T(\F,\V)$ such that $t_2 \to_\R^* t$ and $t_3 \to_\R^* t$. \R is said to be \emph{canonical} if the binary one-step rewriting relation $\to_\R$ is terminating and confluent.

We have defined the lexical, the syntax and the semantic of a simple functional language with evaluation strategy eager without sorts called. The syntax is similar to \textsf{Maude} \cite{clavel00} for specifying arithmetic operations and \textsf{Prolog} \cite{iranzo07} for the notation of list. The language has as constructors the value \texttt{0} (cardinal of the empty set), the unary function \texttt{s} successor ($\mathtt{s}:\mathbb{N} \to \mathbb{N}^+: n \mapsto \mathtt{s}(n) = n + 1$), the notation for list is $\texttt{[}H\texttt{|}T\texttt{]}$ where $H$ is the head of the list and $T$ is the tail of the list, the empty list is denotes by the string \texttt{[]}, into the system a list is denoted as the binary function $\bullet(H,T)$ (dot operator), the name of the functions are non-empty sequences of small letters and the variables are non-empty sequences of capital letters.

\section{Genetic Programming (GP)}

Genetic Programming is a branch of Genetic Algorithms, in which individuals (chromosomes) of population are computer programs \cite{Koza92,Koza94,Koza99,Koza03}. The main difference between GP and traditional genetic algorithms is that the representation of the solution. GP creates computer programs in some particular programming language to represent a solution, whereas genetic algorithms create a string of numbers that represent the solution. The goal of the GP is to find a program that solves a problem such that its analytic solution is very complicated to find. The fitness of a valid individual is generally obtained by the performance and the behavior of it over a training data sets. As any program can be represented as a tree or a set of trees (genotype), the programs (chromosomes) usually are represented as data structures (trees), these trees are obtained doing parsing over sentences (strings) of a program (phenotype), from where GP is applied over a particular domain (programming language). The fitness of a valid individual is generally obtained by the performance and the behavior of it over a training data sets.

\subsection{Functions and Terminals in GP}

The set of terminals and functions is the most important component of the GP. The set of terminals and functions is the alphabet of the programs that are build over the language programming. The variables and constants of the programs belong to set of terminals.

\subsection{Representation of programs} 

As any program can be represented as a tree or a set of trees (genotype), the programs (chromosomes) usually are represented as data structures (trees), these trees are obtained doing parsing over sentences (strings) of a program (phenotype), from where GP is applied over a particular domain (programming language).

\subsection{Generation of initial populations}

As the mutation is almost always lethal, it is very restricted, then, the diversity in the generations is obtained only into the initial population, the parameter to build individuals is the maximum depth of the branches of the trees, represented by $l$. The main methods are:
\begin{description}
\item[Full:] Length of all branches equal to $l > 0$.
\item[Grow:] Length of all branches of depth less than or equal to $l > 0$.
\item[Ramped (half-and-half):] The individuals are created with the full or grow methods where trees are of iterative depth $1,2,3,\ldots,l$.  
\end{description} 

\subsection{Genetic Programming Operators}
\begin{description}
\item[Crossover:] given a couple of programs, a subtree of each program is randomly selected and these subtrees are swapped.
\item[Mutation:] given a program, a subtree of this program is randomly selected and it is replaced by a new randomly generated tree.
\end{description}

\section{Hybrid Adaptive Evolutionary\\Algorithm (\haea)}

Parameter adaptation techniques tried to eliminate the parameter setting  process  by  adapting  parameters  through the algorithm's execution.  Parameter adaptation techniques can be roughly divided into centralized control techniques (central learning rule), decentralized control techniques,  and  hybrid  control  techniques.  In  the  centralized  learning  rule  approach, genetic  operator rates  (such  as mutation  rate,  crossover  rate,  etc.)  are  adapted  according to a  global  learning  rule  that  takes  into  account  the  operator productivities through generations (iterations). Generally, only one operator is applied per generation, and it is selected based  on  its  productivity.  The  productivity  of  an  operator is  measured  in  terms  of  good  individuals  produced  by  the operator. A good individual is one that improves the fitness measure of the current population. If an operator generates a higher number of good individuals than other operators then its probability is rewarded in another case this one is punished. 

In decentralized control strategies, genetic operator rates are encoded in the individual and are subject to the evolutionary process. Accordingly, genetic operator rates can be encoded as an array of real values in the semi-open interval [0.0,1.0), with the constraint that the sum of these values must be equal to one. Since the operator rates are encoded as real  numbers,  special  genetic  operators,  meta-operators,  are
applied to adapt or evolve them.

In \cite{Gomez04a,Gomez04b} was proposed an evolutionary algorithm that adapts the operator rates while it is solving the optimization problem. The  Hybrid  Adaptive  Evolutionary  Algorithm~(\haea)  is a  mixture  of  ideas  borrowed  from  Evolutionary  Strategies,  decentralized  control  adaptation,  and  central  control
adaptation. 

In \haea, each individual is ``independently'' evolved from the  other  individuals  of  the  population,  as  in  evolutionary strategies.  In  each  generation,  every  individual  selects only  one  operator  from  the  set  of  possible  operators. Such operator is selected according to the operator rates encoded  into  the  individual.  When a non-unary operator  is applied,  additional  parents  (the  individual  being  evolved  is considered  a  parent)  are  chosen  according  to  any  selection strategy.  As  can  be  noticed,  \haea   does  not generate a parent population from which the next generation is totally produced. Among the offspring produced by the genetic operator, only one individual is chosen as child, and
will take the place of its parent in the next population. In order to be able to preserve good individuals through evolution, \haea  compares the parent individual against the offspring generated by the operator. The best selection mechanism will determine the individual (parent or offspring) that
has the highest fitness. Therefore, an individual is preserved through evolution if it is better than all the possible individuals generated by applying the genetic operators. 

The genetic operator rates are encoded into the individual in the same way as decentralized control adaptation.  These  probabilities  are  initialized with values following a standard uniform distribution $U(0,1)$. A roulette selection scheme is used to select the operator to be applied. To do this, the operator rates  are normalized in  such a way  that their summation is equal to one.

The performance of the child is compared against its parent performance  in  order  to  determine  the  productivity  of  the operator. The operator is rewarded if the child is  better  than  the  parent  and  punished  if  it  is  worst.  The magnitude of reward/punishment is defined by a learning rate that  is  randomly  generated.  Finally,  operator  rates are  recomputed, normalized,  and  assigned  to  the  individual that will be copied to the next population. The learning rate is generated in a random fashion instead of setting it to a specific value.


\section{Genetic Inductive Functional\\Programming}

The evolutionary algorithm proposed takes as entry a finite set of concrete examples (pairs input/output) of the function (functional program) to induce, the dataset is divided into two sets, the first is the positive basic examples $E^+$ (or the training set), second is the positive extra examples $E^{++}$ (or the validation set) and optionality a program as background knowledge, to continuation we present the evolutionary algorithm proposed.

\subsection{Representation of Individuals}

In our evolutionary algorithm individuals or genomes are ordered lists of trees, these lists are in genotype space and the list of equations represented by these trees are in the phenotype space. 

\begin{example} Following list of equations represents an individual of the evolutionary algorithm.
\begin{verbatim}
sum(N,0) = N
sum(N,s(M)) = s(sum(N,M))
prod(N,0) = 0
prod(N,s(M)) = sum(prod(N,M),N)
double(0) = 0
double(s(N)) = s(s(double(N)))
triple(0) = 0
triple(s(N)) = s(s(s(triple(N))))
square(0) = 0
square(s(N)) = sum(square(N),sum(s(N),N))
cube(0) = 0
cube(s(N)) = s(sum(cube(N),triple(sum(square(N),N))))
\end{verbatim}
\end{example}   

\subsection{Generation of the Initial Population}

From set of concrete examples is possible to obtain an initial population using a process of generalization to each example. Generalization and restricted generalization concepts are defined to continuation   

\begin{definition}(Generalization)
Given a ground term $t$, $s$ is a \emph{generalization} of $t$, if there exists a ground replacement $\sigma$ such that $\sigma(s) \equiv  t$. 
\end{definition}

\begin{definition}(Restricted Generalization)
Given a ground term $t$, $s$ is a \emph{restricted generalization} of $t$, if $s$ is a generalization of $t$, and $s[\Lambda] \equiv \texttt{=}$, i.e. $s\equiv X = Y $, $s[1]$ is a function non-constant and $\forall x \in \Var(Y), x \in \Var(X)$. 
\end{definition}

Initial population is obtained building the set of all restricted generalizations of each concrete example. The restricted generalization in our algorithm are obtained replacement each subterm of an example by variables in all possible forms. Individuals in our algorithm is composed by two disjoint set, the first to obtain the basic equations (base cases) and the second to obtain the recursive equations, those basic and recursive sets are initially composed by restricted generalizations. 

\begin{example}
Given the example \verb|square_bino(0,0) = 0|, in Table~\ref{tbl:generalization} are presented all generalizations of this equation, after we delete the generalizations such that lhs is not a function or there exists variables in rhs that are not int lhs, in Table~\ref{tbl:restricted:generalization} are presented all restricted generalizations of the equation, this set is used as initial population of the evolutionary algorithm proposed.    
\begin{table}[!htb]
\centering
\begin{tabular}{r|l}\hline
\multicolumn{1}{c|}{\textbf{No.}} & \multicolumn{1}{c}{\textbf{Generalization}} \\\hline
1  & \verb|A = B| \\
2  & \verb|A = 0| \\
3  & \verb|square_bino(A,B) = C| \\
4  & \verb|square_bino(A,B) = B| \\ 
5  & \verb|square_bino(A,B) = A| \\
6  & \verb|square_bino(A,B) = 0| \\
7  & \verb|square_bino(A,A) = B| \\
8  & \verb|square_bino(A,A) = A| \\
9  & \verb|square_bino(A,A) = 0| \\
10 & \verb|square_bino(A,0) = B| \\
11 & \verb|square_bino(A,0) = A| \\
12 & \verb|square_bino(A,0) = 0| \\
13 & \verb|square_bino(0,A) = B| \\
14 & \verb|square_bino(0,A) = A| \\
15 & \verb|square_bino(0,A) = 0| \\
16 & \verb|square_bino(0,0) = A| \\
17 & \verb|square_bino(0,0) = 0| \\\hline
\end{tabular}
\caption{Generalizations of the example\\ \texttt{square\_bino(0,0) = 0}.}\label{tbl:generalization}
\end{table}
\begin{table}[!htb]
\centering
\begin{tabular}{r|l}\hline
\multicolumn{1}{c|}{\textbf{No.}} & \multicolumn{1}{c}{\textbf{Restricted Generalization}} \\\hline
4 &	\verb|square_bino(A,B) = B| \\
5 &	\verb|square_bino(A,B) = A| \\
6 &	\verb|square_bino(A,B) = 0| \\
8 & \verb|square_bino(A,A) = A| \\
9 &	\verb|square_bino(A,A) = 0| \\
11 & \verb|square_bino(A,0) = A| \\
12 & \verb|square_bino(A,0) = 0| \\
14 & \verb|square_bino(0,A) = A| \\
15 & \verb|square_bino(0,A) = 0| \\
17 & \verb|square_bino(0,0) = 0| \\\hline
\end{tabular}
\caption{Restricted generalizations of the example \texttt{square\_bino(0,0) = 0}.}\label{tbl:restricted:generalization}
\end{table}
\end{example}

As the number of restricted generalizations are limited, then the initial population can have duplicate individuals to establish a minimum size of the population. 

\subsection{Genetic Operators}

In this Section we present a set of nine operators, three binaries and six unaries, two of these operators are the classical operators of genetic programming, another operators are built in such a way that these preserve the syntax of programs induced.

\subsubsection{Global XOver Operator}

The global xover operator is a binary operator, that randomly selects an equation from each individual, and exchanges them (at random positions).

\begin{example}
Given two individuals $\{p_1, p_2\}$ defined as follows
\begin{align*}
p_1 &= \{\texttt{sum\_n(N) = N}, \texttt{sum\_n(s(N)) = sum(N,sum\_n(N))}\}\\
p_2 &= \{\texttt{sum\_n(s(N)) = s(sum(N,sum\_n(N)))}, \texttt{sum\_n(0) = 1}\}
\end{align*} 
the global xover operator selects  two equations, in this example, equations $\texttt{sum\_n(s(N)) = sum(N,sum\_n(N))}$ of $p_1$ and\break $\texttt{sum\_n(s(N)) = s(sum(N,sum\_n(N)))}$ of $p_2$, exchanges both of them and obtains the following new individuals 
\begin{align*}
p'_1 &= \{\texttt{sum\_n(s(N)) = s(sum(N,sum\_n(N)))},
\texttt{sum\_n(N) = N},\}\\
p'_2 &= \{\texttt{sum\_n(s(N)) = sum(N,sum\_n(N))}, 
\texttt{sum\_n(0) = 1}\}
\end{align*} 
\end{example}

\subsubsection{Global Swap Operator}

The global swap operator is an unary operator, that takes an individual and swaps two randomly selected equations if it is possible (if the individual has at least two equations). Otherwise, the individual remains the same. 

\begin{example}
Given an individual $p$ defined as follows
\begin{multline*}
p = \{\texttt{sum\_n(N) = N}, \texttt{sum\_n(s(N)) = sum(s(N),sum\_n(N))}\}
\end{multline*}
the global swap operator selects two equations of $p$, in this example, equations $\texttt{sum\_n(s(N)) = sum(s(N),sum\_n(N))}$\break and $\texttt{sum\_n(N) = N}$, and swaps both of them obtaining following new individual.
\begin{multline*}
p' = \{\texttt{sum\_n(s(N)) = sum(s(N),sum\_n(N))}, \texttt{sum\_n(N) = N}\}
\end{multline*} 
\end{example} 

\subsubsection{Internal Swap Operator}

The internal swap operator is an unary operator, that randomly selects an equation, in it selects a random function, and then two different parameters are also randomly selected if it is possible (if the function has arity greater than or equals to two), and these parameters are swapped.  Otherwise, the individual remains the same.

\begin{example}
Given an individual $p$ defined as follows 
\begin{multline*}
p = \{\texttt{prod(N,0) = 0}, \texttt{prod(s(M),N) = sum(N,prod(N,M))}\}
\end{multline*}
the internal swap operator selects the equation\break $\texttt{prod(s(M),N) = sum(N,prod(N,M))}$ of $p$, and in it selects function $\texttt{prod}$ at position $1$, and swaps both parameters (\texttt{s(M)} and \texttt{N}) obtaining following new individual 
\begin{multline*}
p' = \{\texttt{prod(N,0) = 0}, \texttt{prod(N,s(M)) = sum(N,prod(N,M))}\}
\end{multline*} 
\end{example} 

\subsubsection{Equalization Operator}

The equalization operator is a binary operator, that randomly selects an equation from each individual, the first one is called the \emph{receptor equation} and the second one is called the \emph{emitter equation}. Constants on lhs of the receptor equation are replaced by variables of the receptor equation. Variables of the rhs of the receptor equation are replaced by variables on lhs of the receptor equation, and constants follow the same process but including a random element, i.e. replaced with a probability of 0.5. This full process is applied to the emitter equation as well.

 if subterms of the rhs of each equation can unify, then one of this unify term of the rhs of one of those equations is replaced by the lhs of the another equation, if the rhs of the new equation has variables that are not present in the lhs, then these variables are replaced by randomly selected variables from the lhs.  

\begin{example}
Given two individuals $\{p_1, p_2\}$ defined as follows
\begin{align*}
p_1 &= \{\texttt{sum\_n(s(A)) = sum(s(A),A)}\} \\
p_2 &= \{\texttt{sum\_n(A) = A}\}
\end{align*}
the equalization operator selects a random equation of each individual \texttt{sum\_n(s(A)) = sum(s(A),A)} of $p_1$ and \texttt{sum\_n(A) = A} of $p_2$, the first one is the receptor equation and the second one is the  equation emitter, obtains equation \texttt{sum\_n(N) = N} replacing in the emitter equation all variables by fresh variables.   Subterms of the rhs  of the receptor equation (\texttt{sum(s(A),A)}, \texttt{s(A)}, \texttt{A}, \texttt{A}) are unifying with subterms of the rhs of the new emitter equation (\texttt{N}) and replaces all subterms of the rhs of the receptor equation by the lhs of the new emitter equation obtaining following new equations
\begin{align*}
\texttt{sum\_n(s(A))} &\texttt{ = sum\_n(N)} \\
\texttt{sum\_n(s(A))} &\texttt{ = sum(sum\_n(N),A)}\\
\texttt{sum\_n(s(A))} &\texttt{ = sum(s(sum\_n(N)),A)}\\
\texttt{sum\_n(s(A))} &\texttt{ = sum(s(A),sum\_n(N))}
\end{align*}   

the operator replaces variables of rhs of the new equations by random variables on lhs obtaining following new equations 
\begin{align*}
\texttt{sum\_n(s(A))} &\texttt{ = sum\_n(A)} \\
\texttt{sum\_n(s(A))} &\texttt{ = sum(sum\_n(A),A)}\\
\texttt{sum\_n(s(A))} &\texttt{ = sum(s(sum\_n(A)),A)}\\
\texttt{sum\_n(s(A))} &\texttt{ = sum(s(A),sum\_n(A))}
\end{align*}
builds a new individual joins equations of initial individuals and randomly inserts last new equations in this new individual obtaining following new individuals
\begin{multline*}
p'_1 = \{\texttt{sum\_n(s(A)) = sum(s(A),A)}, \texttt{sum\_n(A) = A}, \\
\texttt{sum\_n(s(A)) = sum\_n(A)}\} 
\end{multline*}
\begin{multline*}
p'_2 = \{\texttt{sum\_n(s(A)) = sum(s(A),A)}, \texttt{sum\_n(A) = A}, \\
\texttt{sum\_n(s(A)) = sum(sum\_n(A),A)}\} 
\end{multline*}
\begin{multline*}
p'_3 = \{\texttt{sum\_n(s(A)) = sum(s(A),A)}, \texttt{sum\_n(A) = A}, \\
\texttt{sum\_n(s(A)) = sum(s(sum\_n(A)),A)}\} 
\end{multline*}
\begin{multline*}
p'_4 = \{\texttt{sum\_n(s(A)) = sum(s(A),A)}, \texttt{sum\_n(A) = A}, \\
\texttt{sum\_n(s(A)) = sum(s(A),sum\_n(A))}\} 
\end{multline*}
This operator can generate recursive equations of new individuals.
\end{example}

\subsubsection{Composition Operator}

Composition operator is a binary operator. This operator works as equalization operator, the difference is that one of the programs must be the background knowledge. 

\subsubsection{Functional Swap Operator}

The functional swap operator is an unary operator, that randomly selects an equation, in it selects a random function symbol ($f'$) different to function symbol at position $1$ if it is possible (if the equation has at least two functional calling), and searches the set of function symbols different to function symbol at position $1$ with the same arity of the function symbol $f'$, if the set is not empty, the operator selects a new random function symbol ($f''$) of the set and swaps function symbols $f'$ and $f''$. Otherwise, the individual remains the same.

\begin{example}
Given an individual $p$ defined as follows
\begin{multline*}
p = \{\texttt{prod(N,0) = 0}, \texttt{prod(s(M),N) = prod(N,sum(N,M))}\}
\end{multline*}
the functional swap operator can select equation\break $\texttt{prod(s(M),N) = prod(N,sum(N,M))}$ of $p$, in it selects function symbol $\texttt{sum}$ at position $2.2$, obtains the set $\{\texttt{prod}\}$ of function symbols of arity two that are different of the function symbol at position $1$, into this set selects the function symbol \texttt{prod}, and swaps function symbols $\texttt{sum}$ and $\texttt{prod}$ obtaining following new individual 
\begin{multline*}
p' = \{\texttt{prod(N,0) = 0}, \texttt{prod(s(M),N) = sum(N,prod(N,M))}\}
\end{multline*} 
\end{example} 

\subsubsection{Functional Rename Operator}

The functional rename operator is an unary operator, that randomly selects an equation, in it selects a random function, if the arity of this one is greater than two and there is an unary function as parameter of the function selected, the operator selects a parameter different to the unary function and maps this function to another parameter and replaces the unary function by its single parameter. Otherwise, the individual remains the same.
\begin{example}
Given an individual $p$ defined as follows
$$
p = \{\texttt{sum(N,0) = N},\ \texttt{sum(s(N),M) = s(sum(N,M))}\}
$$
the functional rename operator can select equation\break $\texttt{sum(s(N),M) = s(sum(N,M))}$ of $p$, in it selects function symbol $\texttt{sum}$ at position $1$, since the successor function is unary, maps to the another parameter the successor function and replaces this successor function by its parameter (\texttt{N}) obtaining following new individual
$$
p' = \{\texttt{sum(N,0) = N},\ \texttt{sum(N,s(M)) = s(sum(N,M))}\}
$$
\end{example} 

\subsubsection{XOver GP Operator}

\subsubsection{Mutation GP Operator}

\subsection{Selection of Parents}

In our evolutionary algorithm we use the hybrid adaptive evolutionary algorithm (\haea) that evolves the solutions, this algorithm in each iteration evolves all individuals of the population, and each evolution selects one operator for each individual; if this operator uses another individual to evolve, then randomly are selected four individuals from the population and a tournament selection mechanism is performance with them and winner of this process is selected as the additional individual to use with the operator.   

\subsection{Replacement Strategy}

To obtain next population in our evolutionary algorithm we following a steady state strategy thus, if any child is better than his/her parent, this child replacement the parent in next population and the operator that generates the child is rewarded, else if parent is better than or equal to child then parent is added to the next population but the operator that generates children is punished.

\subsection{Fitness Function}

Fitness of individuals is calculated using the covering of these individuals, it is defined as follows

\begin{definition}(Covering) The \emph{covering} ($Cov$) of a individual (program) $p$ is defined as the set of positive basic and positive extra examples that $p$ can deduce, in others words 
$$
Cov(p)=\big\{e\in E^+ \cup E^{++} : p \models e\big\},
$$ 
and the \emph{covering factor} ($CovF^+$) is defined as the proportion of positive basic and positive extra examples that $p$ can deduce, i.e.
$$
CovF^+(p) = \frac{|Cov(p)|}{|E^+ \cup E^{++}|}
$$
\end{definition} 

Covering factor is a function from the set of programs to the interval $[0,1]$.

From previous definitions fitness function of an individual $p$ is calculated as the covering factor of $p$, and therefore the goal of the evolutionary algorithm is to maximize that function.  

\section{Experiments and Results}

Using the evolutionary algorithm previously described, we try to solve seven problems that consist in to obtain algebraic expressions from arithmetic concrete examples given in form of pairs input/output,    
Our dataset is divided into two sets, the first is the positive basic examples (or the training set) and second is the positive extra examples (or the validation set). 

\subsection{Global Configuration}

In Table~\ref{global:configuration:haea} are presented the parameters used to performance the evolutionary algorithm, as some individuals could be non-terminant programs, it is necessary to limit the number of nodes of each equations, the number of rewrite steps and the number of reducible expression that is searched in each rewrite steps. In Table~\ref{tbl:datasets_bkg} are presented all dataset and the background knowledge of each problem proposed to solve.  

\begin{table}[!htb]
\centering
\renewcommand{\tabcolsep}{4.5pt}
\begin{tabular}{p{70mm}|r<{\ \ }}\hline
\multicolumn{1}{c|}{\textbf{Parameter}} & \multicolumn{1}{c}{\textbf{Value}} \\\hline
Minimum size of the population & 500\\\hline
Number of experiments & 100\\\hline
Maximum number of nodes of each equation & 30\\\hline
Maximum number of rewriting steps & 500\\\hline
Maximum number of searches of reducible expressions & 500\\\hline
Maximum depth of the branches of the individuals  & 2\\
\hline
\end{tabular}
\caption{Configuration proposed of the parameters of the evolutionary algorithm GP.}\label{global:configuration:gp}
\end{table}

\begin{table}[!htb]
\centering
\renewcommand{\tabcolsep}{4.5pt}
\begin{tabular}{p{70mm}|r<{\ \ }}\hline
\multicolumn{1}{c|}{\textbf{Parameter}} & \multicolumn{1}{c}{\textbf{Value}} \\\hline
Minimum size of the population & 500\\\hline
Number of experiments & 100\\\hline
Maximum number of iterations of the \haea algorithm & 100\\\hline
Maximum number of basic equations & 3\\\hline
Maximum number of recursive equations & 3\\\hline
Maximum number of nodes of each equation & 30\\\hline
Maximum number of rewriting steps & 500\\\hline
Maximum number of searches of reducible expressions & 500\\\hline
\end{tabular}
\caption{Configuration proposed of the parameters of the evolutionary algorithm \haea.}\label{global:configuration:haea}
\end{table}

\begin{table*}[!htb]
\centering
\renewcommand{\tabcolsep}{3pt}
\begin{tabular}{l|l|l|l}\hline
& & & \\[-2.0ex]
 \multicolumn{1}{c|}{\textbf{Description}} & \multicolumn{1}{c|}{\textbf{Positive Basic}} & \multicolumn{1}{c|}{\textbf{Positive Extra}} & \multicolumn{1}{c}{\textbf{Background}} \\
 \multicolumn{1}{c|}{\textbf{Problem}} & \multicolumn{1}{c|}{\textbf{Examples $\boldsymbol{E^{+}}$}} & \multicolumn{1}{c|}{\textbf{Examples  $\boldsymbol{E^{++}}$}} & \multicolumn{1}{c}{\textbf{Knowledge}}
\\\hline\hline
\multirow{3}{*}{\parbox{30mm}{Cube of a binomial}} & \verb|cube_bino(0,0) = 0| & \verb|cube_bino(1,0) = 1|, \verb|cube_bino(0,1) = 1|, & \verb|sum|, \verb|prod|,   \\
& & \verb|cube_bino(1,1) = 8|, \verb|cube_bino(2,0) = 8|, & \verb|triple|, \verb|square|, \\
& & \verb|cube_bino(0,2) = 8| & \verb|cube|\\\hline
\multirow{2}{*}{\parbox{30mm}{Cube of the successor\\of a natural number}} & \verb|cube(0) = 0| & \verb|cube(2) = 8|, \verb|cube(3) = 27| & \verb|sum|, \verb|triple|, \\
& \verb|cube(1) = 1| &  &  \verb|square| \\\hline
\multirow{4}{*}{\parbox{30mm}{Square of a binomial}} & \verb|square_bino(0,0) = 0| & \verb|square_bino(1,0) = 1|, \verb|square_bino(0,1) = 1|, & \verb|sum|, \verb|prod|, \\
& & \verb|square_bino(1,1) = 4|, \verb|square_bino(2,1) = 9|, & \verb|double|, \verb|square| \\
& & \verb|square_bino(2,2) = 16|, \verb|square_bino(3,1) = 16|, & \\
& & \verb|square_bino(2,3) = 25|, \verb|square_bino(3,2) = 25| & \\\hline 
\multirow{2}{*}{\parbox{30mm}{Square of the successor\\of a natural number}} & \verb|square(0) = 0| & \verb|square(2) = 4|, \verb|square(3) = 9|, & \verb|sum|, \verb|prod|,  \\
 & \verb|square(1) = 1| &  \verb|square(4) = 16|, \verb|square(5) = 25| & \verb|triple| \\\hline
\multirow{5}{*}{\parbox{30mm}{Square of a trinomial}} & \verb|square_trino(0,0,0) = 0| & \verb|square_trino(0,1,1) = 4|, \verb|square_trino(1,0,1) = 4|, & \verb|sum|, \verb|prod|,\\
& & \verb|square_trino(1,1,0) = 4|, \verb|square_trino(2,0,0) = 4|, &  \verb|double|, \verb|square| \\
& & \verb|square_trino(0,2,0) = 4|, \verb|square_trino(0,0,2) = 4|, & \\
& & \verb|square_trino(1,1,1) = 9|, \verb|square_trino(2,1,1) = 16|, & \\
& & \verb|square_trino(1,2,1) = 16|, \verb|square_trino(1,1,2) = 16| & \\\hline
\multirow{2}{*}{\parbox{30mm}{Sum of the of the first\\$n$ natural numbers}} & \verb|sum_n(0) = 0| & \verb|sum_n(2) = 3|, \verb|sum_n(3) = 6|, \verb|sum_n(4) = 10| & \verb|sum| \\
& \verb|sum_n(1) = 1| & & \\\hline
\multirow{3}{*}{\parbox{30mm}{Sum of the squares\\of the first $n$ natural\\numbers}} & \verb|sum_n_square(0) = 0| & \verb|sum_n_square(2) = 5|, \verb|sum_n_square(3) = 14|, & \verb|sum|, \verb|double|, \\
& \verb|sum_n_square(1) = 1| & \verb|sum_n_square(4) = 30| & \verb|square| \\
& &  & \\\hline
\end{tabular}
\bigskip
\caption{Descriptions of the functions to induce with those dataset and background knowledge.}\label{tbl:datasets_bkg}
\end{table*}

\begin{table*}[!htb]
\centering
\renewcommand{\tabcolsep}{3pt}
\begin{tabular}{l|l|l}\hline
& & \\[-2.0ex]
\multicolumn{1}{c|}{\textbf{Description}} & \multicolumn{1}{c|}{\textbf{Solution}} & \multicolumn{1}{c}{\textbf{Equivalent}} \\
 \multicolumn{1}{c|}{\textbf{Problem}} & \multicolumn{1}{c|}{\textbf{(Program)}}  & \multicolumn{1}{c}{\textbf{Mathematical Expression}}\\ \hline\hline
\multirow{6}{*}{\parbox{18mm}{Cube of a\\binomial}}& & \\[-3mm]
 & \verb|cube_bino(A,B) = cube(sum(A,B))| &
$cube\_bino(A,B) = (A+B)^3$ 
\\\cline{2-3}
& & \\[-2.0ex]
& \begin{minipage}{90mm}
\verb|cube_bino(A,B) =|\\ \verb|sum(prod(sum(sum(prod(A,A),prod(B,B)),|\\
\hspace*{4cm}\verb|sum(prod(B,A),prod(A,B))),B),|\\ \verb|prod(A,sum(sum(prod(B,A),prod(B,A)),|\\
\hspace*{4cm}\verb|sum(prod(B,B),prod(A,A)))))|
\end{minipage} & 
\begin{minipage}{55mm}
$cube\_bino(A,B) =$\\ 
$((AA+BB)\ +$\\ 
$(BA+AB))B\ +$\\ 
$A((BA+BA)\ +$\\ 
$(BB+AA))$ 
\end{minipage}
\\\hline
\multirow{5}{*}{\parbox{18mm}{Cube of the\\successor\\of a natural\\number}} & & \\[-2ex]
& \begin{minipage}{90mm}\verb!cube(0) = 0;!\\ \verb!cube(s(A)) = sum(triple(sum(square(A),A)),s(cube(A)))!\end{minipage} &
\begin{minipage}{55mm}$cube(0) = 0;$\\$cube(A+1) = 3(A^2 + A) + (A^3 + 1)$
\end{minipage}  \\\cline{2-3}
& & \\[-2ex]
& \begin{minipage}{90mm}
\verb!cube(s(A)) =!\\ \verb!sum(sum(sum(triple(square(A)),s(A)),sum(A,cube(A))),A);!\\
\verb!cube(A) = A!
\end{minipage} &
\begin{minipage}{55mm}
$cube(A+1) =$\\$\big(\big(3A^2 + (A+1)\big) + \big(A + A^3\big)\big)+A$;\\ $cube(A) = A$
\end{minipage}\\\hline
\multirow{3}{*}{\parbox{18mm}{Square of a\\binomial}} & & \\[-2.1ex]
&
\verb!square_bino(A,B) = square(sum(B,A))!
&  $square\_bino(A,B) = (B+A)^2$
\\\cline{2-3}
& & \\[-2.1ex]
& \begin{minipage}{90mm}
\verb|square_bino(A,B) =|\\ \verb|sum(sum(prod(A,A),double(prod(A,B))),prod(B,B))|
\end{minipage} &
\begin{minipage}{55mm}
$square\_bino(A,B) = (AA+2AB)+BB$
\end{minipage}
\\\hline
\multirow{5}{*}{\parbox{18mm}{Square of the\\successor\\of a natural\\number}} & & \\[-2.1ex]
 & 
\begin{minipage}{90mm}
\verb!square(s(A)) = sum(square(A),s(double(A)))!;\\ \verb!square(0) = 0!
\end{minipage} & 
\begin{minipage}{55mm}
$square(A+1) = A^2 + (2A+1)$;\\ $square(0) = 0$
\end{minipage}\\\cline{2-3}
& & \\[-2.1ex]
& \begin{minipage}{90mm}
\verb!square(0) = 0;!\\ 
\verb!square(s(A)) = sum(sum(A,square(A)),s(A))!
\end{minipage} & 
\begin{minipage}{55mm}
$square(0) = 0;$\\ $square(A+1) = \big(A + A^2\big) + (A+1)$
\end{minipage}\\\hline
\multirow{3}{*}{\parbox{18mm}{Square of a\\Trinomial}} & & \\[-2.1ex]
& \begin{minipage}{90mm}
\verb|square_trino(A,B,C) = square(sum(B,sum(C,A)))|
\end{minipage} &
\begin{minipage}{55mm}
$square\_trino(A,B,C) = (B + (C+A))^2$
\end{minipage}\\\cline{2-3}
& & \\[-2.0ex]
& \begin{minipage}{90mm}
\verb|square_trino(A,B,C) =|\\ \verb|sum(prod(sum(C,A),sum(B,sum(sum(B,C),A))),prod(B,B))|
\end{minipage} &
\begin{minipage}{55mm}
$square\_trino(A,B,C) =$\\ $(C+A)(B+((B+C)+A))+BB$
\end{minipage}\\\hline
\multirow{6}{*}{\parbox{18mm}{Sum of the\\of the first\\$n$ natural\\numbers}} & & \\[-2.1ex]
& \begin{minipage}{90mm}
\verb|sum_n(0) = 0; sum_n(s(A)) = s(sum(sum_n(A),A))|
\end{minipage} &
\begin{minipage}{55mm}
$\sum_{n=0}^{0} n = 0;$\\ $\sum_{n=0}^{A+1} n =  \big(\sum_{n=0}^{A} n + A\big)+1$
\end{minipage}\\[-2.5ex] & & \\\cline{2-3}
& & \\[-2.0ex]
& \begin{minipage}{90mm}
\verb|sum_n(s(A)) = sum(s(A),sum_n(A)); sum_n(A) = A|
\end{minipage} &
\begin{minipage}{55mm}
$\sum_{n=0}^{A+1} n =  \big((A + 1) + \sum_{n=0}^{A} n\big);$\\$\sum_{n=A}^{A} n = 0$
\end{minipage}\\[-2.5ex]
& & \\\hline
\multirow{6}{*}{\parbox{18mm}{Sum of the\\squares\\of the first\\$n$ natural\\numbers}}
& & \\[-2.1ex]
& \begin{minipage}{90mm}
\verb!sum_n_square(s(A)) =!\\ \verb!sum(sum(square(A),A),sum(s(sum_n_square(A)),A));!\\
\verb!sum_n_square(A) = A!
\end{minipage} &
\begin{minipage}{55mm}
$\sum_{n=0}^{A+1} n^2 = \big(A^2 + A\big) + \big(\big(\sum_{n=0}^{A} n^2 +1\big)+A\big);$\\ $\sum_{n=A}^{A} n^2 = 0$
\end{minipage}\\[-2.5ex] & & \\\cline{2-3}
& & \\[-2.0ex]
& \begin{minipage}{90mm}
\verb!sum_n_square(s(A)) =!\\ \verb!sum(sum(sum_n_square(A),s(square(A))),sum(A,A));!\\ \verb!sum_n_square(0) = 0!
\end{minipage} &
\begin{minipage}{55mm}
$\sum_{n=0}^{A+1} n^2 = \big(\sum_{n=0}^{A} n^2 + \big(A^2 + 1\big)\big) + (A+A);$\\ $\sum_{n=0}^{0} n^2 = 0$
\end{minipage}\\[-2.5ex]
& & \\\hline
\end{tabular}
\caption{Examples of the programs solutions and those respective algebraic expressions found using the evolutionary algorithm 
from dataset and background knowledge presented in Table~\ref{tbl:datasets_bkg}.}\label{tbl:solutions:expressions}
\end{table*}

\subsection{Global Results}

Using dataset of each problem presented previously, and the configuration of parameters defined in Table~\ref{global:configuration}, we obtained following results, where
each stacked bar chart represents the total of experiments, subarea with diagonal lines represents the proportion of successful experiments and empty area represents the proportion of fail experiments. 

\begin{table}[!htb]
\begin{footnotesize}
\renewcommand{\tabcolsep}{3pt}
\begin{tabular}{l|rrl}\hline\\[-2ex]
\multicolumn{1}{c|}{Problem/Program} & 
\multicolumn{1}{c}{\ding{51}} &
\multicolumn{1}{c}{\ding{55}} &
\multicolumn{1}{c}{Summary} \\\hline
cube-bino & 96 & 4 & \barchart[scale=0.025,experiments=100,success=96,persuccess=96.0,minlength=44]\\
cube & 0 & 100 & \barchart[scale=0.025,experiments=100,success=0,persuccess=0.0,minlength=0]\\
square-bino & 81 & 19 & \barchart[scale=0.025,experiments=100,success=81,persuccess=81.0,minlength=4]\\
square & 0 & 100 & \barchart[scale=0.025,experiments=100,success=0,persuccess=0.0,minlength=0]\\
square-trino & 28 & 72 & \barchart[scale=0.025,experiments=100,success=28,persuccess=28.0,minlength=2]\\
sum-n & 1 & 99 & \barchart[scale=0.025,experiments=100,success=1,persuccess=1.0,minlength=1]\\
sum-n-square & 0 & 100 & \barchart[scale=0.025,experiments=100,success=0,persuccess=0.0,minlength=0]\\
\hline
\end{tabular}
\caption{GP = 7h-2m-35s}
\end{footnotesize}
\end{table}

\begin{table}[!htb]
\renewcommand{\tabcolsep}{3pt}
\begin{footnotesize}
\begin{tabular}{l|rrl}\hline\\[-2ex]
\multicolumn{1}{c|}{Problem/Program} & 
\multicolumn{1}{c}{\ding{51}} &
\multicolumn{1}{c}{\ding{55}} &
\multicolumn{1}{c}{Summary} \\\hline
\hline
cube-bino & 100 & 0 & \barchart[scale=0.025,experiments=100,success=100,persuccess=100.0,minlength=1]\\
cube & 8 & 92 & \barchart[scale=0.025,experiments=100,success=8,persuccess=8.0,minlength=2]\\
square-bino & 100 & 0 & \barchart[scale=0.025,experiments=100,success=100,persuccess=100.0,minlength=82]\\
square & 99 & 1 & \barchart[scale=0.025,experiments=100,success=99,persuccess=99.0,minlength=1]\\
square-trino & 97 & 3 & \barchart[scale=0.025,experiments=100,success=97,persuccess=97.0,minlength=1]\\
sum-n & 100 & 0 & \barchart[scale=0.025,experiments=100,success=100,persuccess=100.0,minlength=100]\\
sum-n-square & 69 & 31 & \barchart[scale=0.025,experiments=100,success=69,persuccess=69.0,minlength=3]\\
\hline
\end{tabular}
\caption{\haea = 10h-56m-54s}
\end{footnotesize}
\end{table}

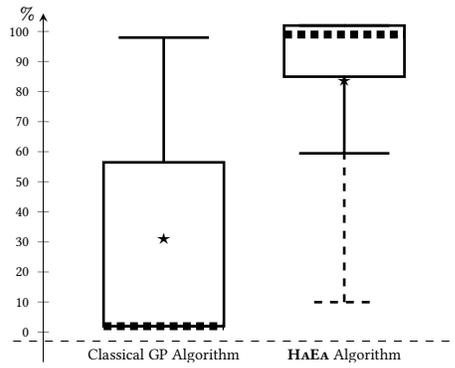
\begin{figure}[htb!]
\centering
\begin{tikzpicture}
\begin{axis}
[x=8mm,y=0.4mm,xmin=-0.5,xmax=7,ymin=-10,ymax=106,
axis y line=middle,axis x line=none, ylabel={$\%$},xtick=\empty,ytick={0,10,...,100},
ticklabel style={font=\tiny},ylabel style={left=0pt},xlabel style={below=3mm}]
\draw[dashed,-stealth] (axis cs:-0.5,-3) -- (axis cs:7,-3);
\end{axis}
\bigboxwhiskers[title=Classical GP Algorithm,numberbox=1,lowerinnerfence=-81.75,min=0,Q1=0.0,median=0.0,Q3=54.5,max=96,upperinnerfence=136.25,mean=29.428571428571427]
\bigboxwhiskers[title=\haea Algorithm,numberbox=2,lowerinnerfence=57.5,min=8,Q1=83.0,median=97.0,Q3=100.0,max=100,upperinnerfence=125.5,mean=81.85714285714286]
\end{tikzpicture}
\caption{Box-and-whisker diagram: .}
\end{figure}

In Table~\ref{tbl:solutions:expressions} are presented some of solutions found using the evolutionary algorithm proposed, each solution is interpreted as mathematical expression, thus those expressions will be equivalents using transitivity property of equality of natural numbers, and so to establish some notable products. 
  
\section{Conclusions}
In this paper we presented an evolutionary algorithm that permit to obtain equivalent algebraic expressions of some notable products and summations, these expressions were obtained using an inductive process from sets of concrete examples (pairs input/output), the algebraic expressions was obtained from arithmetic equalities and not as usually is done from a deductive method from high-level description using  Field Axioms for the real numbers  \cite{apostol2007calculus} as is done usually.     

It is important to mention that solutions found are correct with respect to each dataset, is to said, solution can deduce all examples presented, but also could deduce values that do not belong to desired function to induce, so our evolutionary algorithm is correct with respect to the set of fact presented.

The background knowledge is a factor very important in the induction of interesting programs, since if the background knowledge contains a lot information (functions) the algorithm prefers to obtain sort solutions, and in this case those kind of solution are the trivial solutions.   

The quality and quantity of the examples presented determine the expression obtained, since if to the algorithm is presented few examples, it could find solutions faster, but these solutions usually are not the solution that we desired. For example, if in the problem of the cube of a successor of a natural number is deleted the example \verb|cube_succe(4) = 64| the solution found is incorrect, on the contrary, if a lot examples are added then to find a solution is very hard.    

The dataset presented in Table~\ref{tbl:datasets_bkg} could be used as benchmark functions to prove another algorithms that are used to obtain representation of theories as algebraic expressions, since we prove that with these dataset and background knowledges it is possible to generate the function that explains the dataset with algebraic expressions.    

\bibliographystyle{abbrv}
\bibliography{references}  

\bibliographystyle{ACM-Reference-Format}
\bibliography{references} 

\end{document}